\documentclass{IEEEtran} 
\usepackage{graphicx} 

\usepackage{bm}
\usepackage{mathrsfs}
\usepackage{array}
\usepackage{multirow}
\usepackage{threeparttable}
\usepackage{xcolor}
\usepackage{amsmath}
\usepackage{url}
\usepackage[final]{changes}
\let\proglang=\textsf

\title{\LARGE \bf
\added{News and Load: A Quantitative Exploration of Natural Language Processing Applications for Forecasting Day-ahead Electricity System Demand}
}
\author{Yun Bai, \textit{Member, IEEE}, Simon Camal, \textit{Member, IEEE}, Andrea Michiorri
\thanks{The authors are with the Centre for Processes, Renewable Energies and Energy Systems (PERSEE), MINES Paris - PSL University, Sophia Antipolis, France, e-mail: (name.surname@minesparis.psl.eu). The author Yun Bai was supported by the program of the China Scholarship Council (CSC Nos. 202106020064).}%
}
\begin{document}

\maketitle
\thispagestyle{empty}
\pagestyle{empty}

\begin{abstract}
The relationship between electricity demand and weather is well established in power systems, along with the importance of behavioral and social aspects such as holidays and significant events. 
This study explores the link between electricity demand and more nuanced information about social events. 
This is done using mature Natural Language Processing (NLP) and demand forecasting techniques. 
The results indicate that day-ahead forecasts are improved by textual features such as word frequencies, public sentiments, topic distributions, and word embeddings. 
The social events contained in these features include global pandemics, politics, international conflicts, transportation, etc. 
Causality effects and correlations are discussed to propose explanations for the mechanisms behind the links highlighted. 
This study is believed to bring a new perspective to traditional electricity demand analysis. 
It confirms the feasibility of improving forecasts from unstructured text, with potential consequences for sociology and economics.
\end{abstract}


\section{Introduction}\label{Introduction}
Load demand forecasting has become integral to the electricity market decision-making process. 
In practice, accurate forecasting assists operators in efficiently scheduling resources and detecting vulnerable situations. 
It also works as a reference for participants to maximize their benefits and improve demand-side management in smart grids \cite{metaxiotis2003artificial}.
Earlier research has identified that meteorology (e.g., temperature) and human activity (e.g., weekends and weekdays) have a crucial impact on load forecasting, and models have been developed incorporating these external features \cite{juberias1999new}.
\added{
According to the analysis in \cite{zhang2019novel} and \cite{petropoulos2022forecasting}, seasonal temperatures impact household load consumption by increasing or decreasing the usage of air conditioners, fans, and heating devices. 
The influence can be found by observing the positive or negative correlations between temperatures and load in different months. 
The correlations assist in forecasting procedures, especially when the autoregression effects of historical load are decreasing \cite{sgarlato2022role}.
}

Recently, social and economic aspects have been recognized to play an important role in variations in electricity demand. vv
Global events, such as pandemics, climate change, and international conflicts, have challenged previous assumptions and created unseen electricity market conditions.
The textual information present in public news constitutes a tangible representation of societies.
In this context, the motivation behind this research is to extract valuable knowledge from these sources, to verify how to establish a link with electricity demand, and to propose explanations. 
A practical application is to improve demand forecasting to increase the ability of power systems to withstand uncertain events.

Utilizing the information available in text has undoubted potential for demand forecasting, although it is subject to problems arising from the inherent characteristics of text. 
Text is typically unstructured, with no predefined format or organization, and is difficult to collect, process, and analyze \cite{pickell2018structured}. 
The field of text-based forecasting is relatively new but has been explored by researchers, with a visible acceleration after 2010 and a general interest in price predictions.
The first attempt can be traced in \cite{joshi2010movie}, where the fundamental concepts of text-based forecasting were suggested, and tests were carried out for forecasting movie revenue using n-grams, part-of-speech n-grams, and the dependency relations from online movie reviews.
Since then, the text-based forecasting approach has been tested in several scenarios, such as bankruptcy and fraud \cite{cecchini2010making}, stock prices \cite{schumaker2010discrete}, demand for taxi rides \cite{rodrigues2019combining}, COVID-19 evolution \cite{zhang2020predicting}, and tourism demand from online searches \cite{huang2021research}.

In applications, text is converted into numerical values through Natural Language Processing (NLP) methods.
The main NLP techniques include sentiment analysis, topic modeling, and word embedding, etc. 
In \textbf{sentiment analysis}, polarity portrays the sentiment tendency (positive or negative) within the text; subjectivity quantifies the amount of personal opinion conveyed in a sentence; and emotion represents human feelings such as happiness, sadness, and disappointment \cite{bravo2013combining}. 
\textbf{Topic modeling} can identify abstract expressions in text, explain the contents using meaningful keywords, and calculate the topic probability distribution of a given text \cite{papadimitriou1998latent}. 
\textbf{Word embedding} aims at mapping words to vectors in the high dimensional space. 
Word vectors are closer in distance when they are similar in lexical meaning.

The following relevant works show how textual information can be applied in energy and power systems.
In \cite{bai2022crude}, the authors improved the crude oil price forecasts by adding public sentiments and news topic distributions.
The authors in \cite{sun2016data} developed a supervised topic model to identify related topics in Twitter data and assist in detecting power outages. 
The research in \cite{wagy2017crowdsourcing} established the relationship between crowd-sourced data and electricity consumption.
This research considered the social aspects of the electricity system but did not explore the textual information within the contents.
In \cite{obst2021textual}, weather reports and tweets were used to improve electricity demand prediction, and specifically the sudden demand changes caused by COVID-19 lockdowns in France and Italy.
In the following work, the author employed the number of tweets with the word ``télétravail" (French for ``remote working") as a feature to improve the benchmark model \cite{obst2021adaptive}.
However, they did not include semantic information in text.
In \cite{wang2023forecasting}, the authors built a Convolutional Neural Network (CNN) model, converted text to word vectors, and fed them as features into a forecasting model for Chinese electricity consumption data.
The above studies do make some progress in text-based forecasting, but lack a discussion of the relationship between forecasting and text, and an insight into how the text mechanism works.

The field of NLP application for forecasting, particularly electricity demand, is at its beginnings.  
This work aims to explore the possibilities of this approach, considering the hypotheses that i) electricity demand is influenced by social factors visible in the news, and ii) NLP enables the quantification of text and its practical use. 
The contributions of this paper are summarized as follows:
\begin{enumerate}
    \item We verified that information extracted from news can improve electricity demand forecasting.
    \item We developed a complete forecasting chain integrating text and other structured data.
    \item We explained the mechanisms behind the improved performance from global, local, and causality perspectives.
\end{enumerate}

The rest of the paper is structured as follows: Section~\ref{Methodology} presents the methodology with an overview of the NLP and forecasting techniques.
The results are presented in Section~\ref{Results} followed by a discussion in Section~\ref{Discussion}, and conclusions are drawn in Section~\ref{Conclusions}.

\section{Methodology}\label{Methodology}
\begin{figure}[ht]
\centerline{\includegraphics[scale=.43]{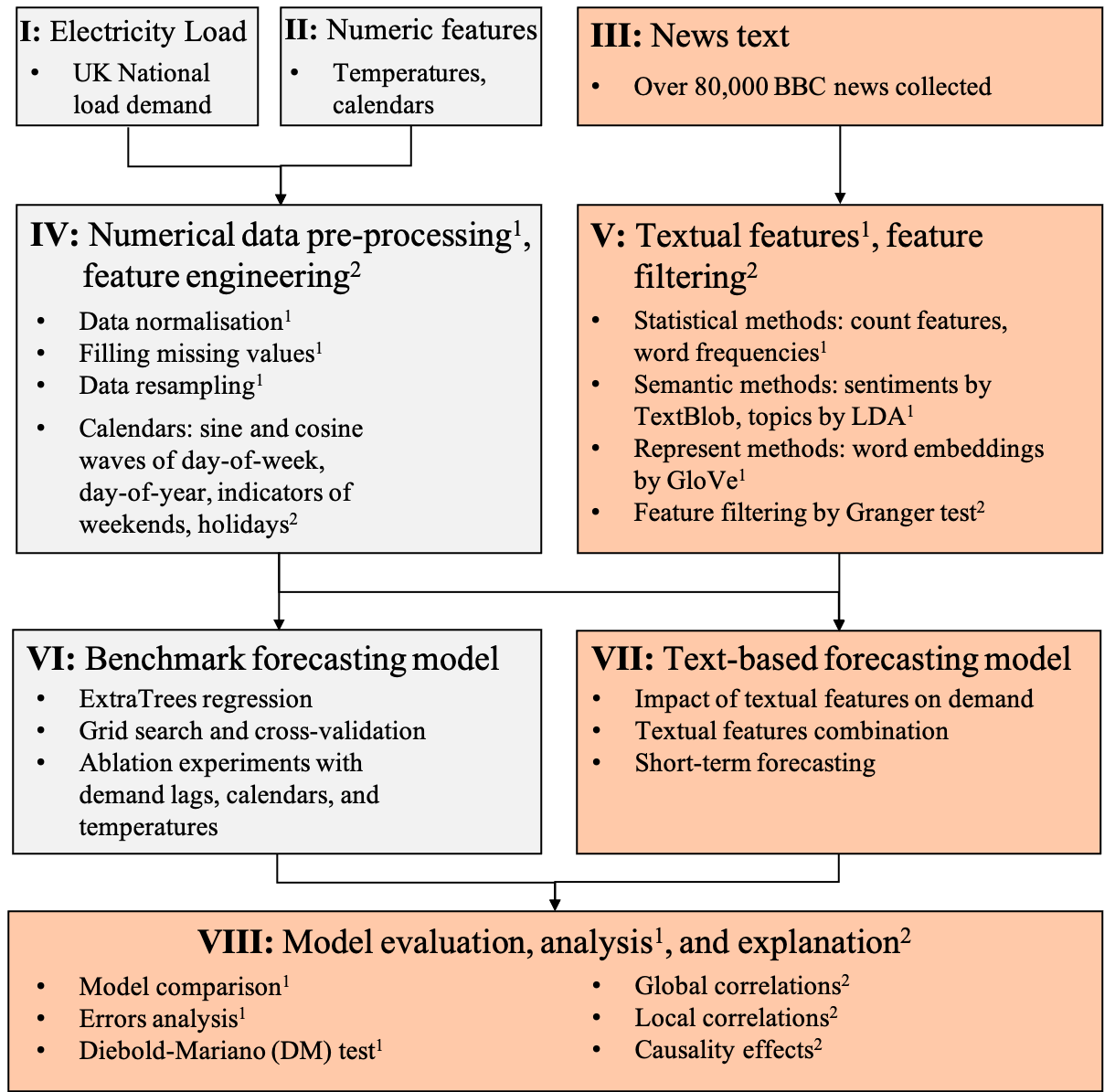}}
\caption{The forecasting framework of this research.}
\label{workflow}
\end{figure}

An overview of the workflow is described in Figure~\ref{workflow}.
Firstly, electricity load, temperatures and calendar data, and news text are acquired in modules \textbf{I}, \textbf{II}, and \textbf{III}.
Then numerical and textual data are pre-processed in modules \textbf{IV} and \textbf{V}.
The process of numerical data includes data normalization, filling missing values, and data resampling.
For calendar data, the features of day-of-week and day-of-year, and indicators of weekends and holidays are considered.
For text data, the inputs are pre-processed to obtain count features and word frequencies with statistical methods; sentiments and topic distributions with semantic methods; and word embeddings with representation methods.
These methods and features will be shown in Section~\ref{Preprocessing for the text-based model}.
Note that some features are redundant for forecasting, and they are filtered out by a Granger test. 
The benchmark model is built in module \textbf{VI} with the numeric features of demand lags, calendars, and temperatures.
Module \textbf{VII} explores adding textual features to the benchmark model. 
Feature combinations are considered to improve forecasting further, and the scenario of short-term forecasting is also tested at this stage.
The final module \textbf{VIII} evaluates the text-based model by comparing it with the benchmark and official forecasts.

\subsection{Pre-processing for the text-based model}
\label{Preprocessing for the text-based model}
This section describes the methods in module \textbf{V} in Figure~\ref{workflow}.
\added{Some pre-processing steps are necessary to remove irrelevant terms, aggregate contents, and increase semantic information \cite{hickman2022text}.}
These steps include splitting a sentence into words, converting all letters to lowercase, and removing stopwords (from a list provided by the Python package Natural Language ToolKit \proglang{NLTK 3.5} \cite{NLTK-web}), words with fewer than three letters, and all numbers \added{which are considered as irrelevant information and noise}.
Each piece of news is then transformed into word lists. 
The NLP techniques for extracting textual features are described below, summarized into statistical, semantic, and representative methods.

\subsubsection{Statistical methods}
\textbf{count features} include 27 daily feature series. 
For each text, we calculated the number of words, sentences, unique words, non-stopping words, average number of sentences, and average number of words for all of the sentences used each day. 
In addition, we made two categorical features corresponding to the proportion of news in the 18 sections of the news website (e.g., Asia, Business, UK Politics).

\textbf{Word frequencies} calculate the number of times that words appear in the news as daily features.
The following word selection criteria are considered: i) stop-words or non-words are not included; ii) only meaningful words with frequencies above a certain threshold can be selected, otherwise the frequency vector will be very sparse and unable to be used for forecasting. 
With the volume of words around 500, the selected words appear enough times, and the frequency vectors are dense.
In total, there are 16, 1, and 0.6 million words in the news bodies (\emph{\textbf{B}}), descriptions (\emph{\textbf{D}}), and titles (\emph{\textbf{T}}), respectively.
If the threshold for the titles is set at  $\sigma_T = 200$, the words that appear fewer than 200 times will be removed from \emph{\textbf{T}}. 
There will be 456 words selected, which is close to 500.
Similarly, we set the thresholds for \emph{\textbf{D}} and \emph{\textbf{B}} $\sigma_D = 400$, $\sigma_B = 5000$, resulting in 329 and 550 words selected.

\subsubsection{Semantic methods}
\label{TextBlob}
\textbf{sentiment analysis} is performed with the library \proglang{TextBlob} from the \proglang{NLTK} package, which is widely used in sentiment analysis and particularly suitable for corpora without manual labeling \cite{kaur2020twitter}.
This method calculates a polarity score between [-1, 1] and a subjectivity score between [0, 1] for each word. 
Then the scores are calculated for each piece of news by averaging the scores from the words. 
We discretized the scores into five buckets and also computed the maximum, minimum, average, and standard deviation for all of the sentiment features.

\textbf{Topic distribution} is analyzed using Latent Dirichlet Allocation (LDA) to obtain a probabilistic estimation that each piece of news belongs to a specific topic \cite{blei2003latent}. 
The daily topic distribution is obtained by averaging the topic probability for all the news during one day, and the probability value under each topic reflects how widespread the daily news is.

\begin{figure}[ht]
    \centering
    \includegraphics[scale=.23]{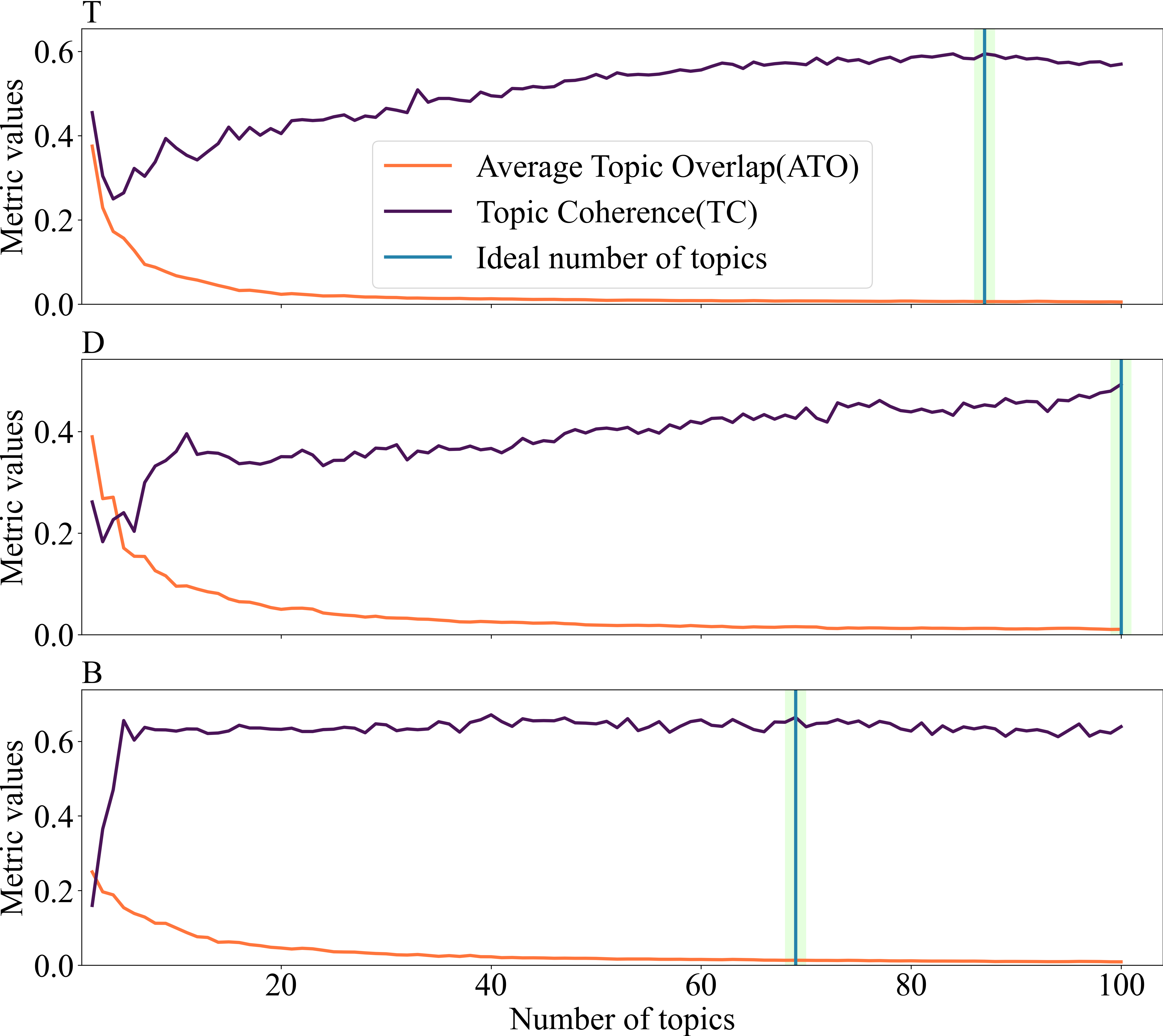}
    \caption{Difference between TC and ATO for $\kappa$s from 1 to 100. The top image represents \emph{\textbf{T}}, the middle one represents \emph{\textbf{D}}, and the bottom one \emph{\textbf{B}}. The vertical lines in each subplot correspond to the ideal $\kappa$s: $\kappa_T=87$, $\kappa_D=100$, $\kappa_B=69$.}
    \label{Difference of topic coherence and average topic overlap}
\end{figure}

The number of topics $\kappa$ in the LDA model should initially be set to summarize the text content better.
The difference between metrics Topic Coherence (TC) and Average Topic Overlap (ATO) can be used to measure whether $\kappa$ is the correct number of topics, as introduced in \cite{li2022exploring}.
TC measures the relevance and coherence of the words within one topic by calculating the similarity of the word pairs in a specific topic \cite{newman2010automatic}.
ATO can be used to evaluate the similarity and overlap between topics under different $\kappa$s \cite{o2015analysis}.
The metric in \cite{li2022exploring} suggests higher TC and lower ATO values.
Thus we plotted the TC and ATO values for $\kappa$s from 1 to 100 in \emph{\textbf{T}}, \emph{\textbf{D}}, and \emph{\textbf{B}} in Figure~\ref{Difference of topic coherence and average topic overlap}.
The most significant difference of TC and ATO for $\kappa$s can be seen as the gap between the two lines in Figure~\ref{Difference of topic coherence and average topic overlap}.

\subsubsection{Representative methods}
This step involves obtaining a vectorial representation of the text through the library \proglang{GloVe} described in \cite{pennington2014glove}. 
This paper utilized Stanford's GloVe 100d word embeddings trained on Wikipedia 2014 and Gigaword 5 (6B tokens, 400K vocab, uncased, 100d vectors).
The pre-trained GloVe not only contains rich semantic information but also facilitates our application on a new corpus without repeating the time-consuming training work.
High-dimensional textual features are extracted and expressed in this work from the textual dataset by transforming words into 100-dimensional vectors with GloVe. 
A text vector is obtained by averaging all the word vectors in this text. 
The position of each element in the text vector is an axis in the high-dimensional space. 
We averaged all the text vectors during one day to obtain the daily embedding features.

\subsubsection{Granger test}
The Granger test, as in module \textbf{V} shown in Figure~\ref{workflow}, is a measure to test whether a stationary time series \emph{X} contributes to the forecasting of \emph{Y} 
\cite{granger1969investigating}.
It is based on the following autoregressive model:
\begin{equation}
\begin{aligned}
    y_t=\theta_0+\sum_{i=1}^{T}\theta_iy_{t-i}&+\sum_{i=1}^T\phi_ix_{t-i}+\epsilon_t\\
    E(\epsilon)&=0,
\end{aligned}
\end{equation}

where $\theta_i$ and $\phi_i$ are the lag coefficients of \emph{X} and \emph{Y}, and \emph{T} is a chosen lag order.
The null hypothesis is 
\begin{equation}
    H_0: \phi_1=\phi_2=...=\phi_T=0,
\end{equation}
i.e. that the lagged terms of \emph{X} are independent of \emph{Y}. 
This can be rejected when $p<0.05$.

Theoretically, the Granger test is a powerful tool that identifies whether a stationary time series can help forecast another one.
For instance, \cite{narayan2005electricity} analyzed the relationship and obtained insights from electricity consumption, employment, and actual income in Australia.
In the application of text-based forecasting,
the authors utilized the Granger test to investigate how online emotions in social media influence stock prices \cite{zhou2016can}.
The Granger test is employed in our research to explore the relationship between textual features and load demand and serves as a feature-filtering technique.
Standing apart from other studies, this paper generated a considerable amount of textual features.
To eliminate the noise in the features, it is necessary to reduce the feature dimensions.
One of our research targets is to explain what information featured in text can help forecast load demand, which is aligned with the principle of the Granger test.

We conducted the bilateral Granger test as in \cite{tang2020multi}.
Two tests are included: $X \rightarrow Y$, and $Y \rightarrow X$.
We aimed to find the feature $X$ that Granger causes demand $Y$ but to avoid $Y$ Granger causes $X$ simultaneously. 
Thus, the results were retained, with $p_1<0.05$ for the first test and $p_2\geq0.05$ for the second one.
Considering that the time series for the Granger test should be stationary, we first carried out the Augmented Dickey-Fuller (ADF) test to verify whether a time series is stationary, referring to \cite{dickey1979distribution}.
We differentiated those series that did not pass the ADF test in order to make them stationary and input them into the Granger test.
The lag orders chosen are those that minimize the Akaike Information Criterion (AIC). 
The AIC was evaluated for four different lag orders applied to daily textual features, namely: a day ($lag=1$), a week ($lag=7$), a month ($lag=30$), and three months ($lag=90$).
To simplify the presentation, we have listed the average AIC values in Table~\ref{AIC values for the Granger test with different lags}.

\begin{table}[ht]
\centering
\caption{AIC values for the Granger test with different lags}
\label{AIC values for the Granger test with different lags}
\begin{tabular}{lcccc}
\hline
Textual features     & Lag1   & Lag7   & Lag30           & Lag90  \\ \hline
Word frequencies      & 13.983 & 13.522 & \textbf{13.408} & 13.548 \\
Sentiment             & 10.518 & 10.030 & \textbf{9.900}  & 10.026 \\
Topic distributions   & 5.268  & 4.790  & \textbf{4.688}  & 4.852  \\
GloVe word embeddings & 7.134  & 6.662  & \textbf{6.543}  & 6.683  \\ \hline
\end{tabular}
\end{table}

As shown in Table~\ref{AIC values for the Granger test with different lags}, the Granger tests with 30 lags have the lowest AIC values and selected fewer than 100 features out of more than 2,000 textual features (See Table~\ref{Forecasting errors when adding Ft into model D+C+T}).
Note that the Granger test used for the initial feature filtering is not an actual causality test, as it involves correlations between lagged and predicted values, and correlations do not necessarily lead to causality.
We discuss the causality of text features and load in more detail in Section~\ref{Causality effects}.

\subsection{Forecasting with textual features}
This section describes the day-ahead forecasting method in modules \textbf{VI} and \textbf{VII} shown in Figure~\ref{workflow}, which we employed to verify whether textual-based features measured in day \emph{d-1} can provide additional explanations for the behavior of demand on day \emph{d+1}.
After an initial comparison with different models, the ExtraTrees regression was selected due to its performance and flexibility \cite{geurts2006extremely}.
ExtraTrees regression is an ensemble learning method.
It creates decision trees during training but randomly samples each tree.
The features in the trees are also randomly selected by splitting values, which enables ExtraTrees to achieve faster speed and membership diversity.
Current applications of ExtraTrees have emerged in the field of load forecasting \cite{kim2020peak,porteiro2019short,xiang2022multi}.

In the dataset of this study ${(X_1, Y_1), (X_2, Y_2), ..., (X_n, Y_n)}$, $X_i$ is the input concatenated by lags $X_{l,i}$, calendars $X_{c,i}$, temperatures $X_{w,i}$, and textual features $X_{t,i}$.
$Y_i$ is the day-ahead load demand vector: $Y_i=(y_{1,i},y_{2,i},...,y_{48,i})$, since the demands are half-hourly.
The ExtraTrees model with $M$ decision trees is fitted, and the $m$-th tree has $T_m$ nodes.
The forecasts of $m$ can be expressed:

\begin{equation}
    f_m(X) = \sum_{t=1}^{T_m}c_{mt} \cdot I (X \in R_{mt}),
\end{equation}

where $c_{mt}$ is the prediction by node $t$, $R_{mt}$ is the feature space region of node $t$, and $I(X \in R_{mt})$ is the indicator function that equals 1 when $X$ belongs to the region $R_{mt}$, and 0 otherwise.
Final ensemble forecasts are the average predictions from all of the decision trees:

\begin{equation}
    f(X) = \frac{1}{M}\sum_{m=1}^{M}f_m(X).
\end{equation}

\subsection{Evaluation}
The regression models in modules \textbf{VI} and \textbf{VII} in Figure~\ref{workflow} are evaluated according to Root Mean Square Error (RMSE), Mean Absolute Error (MAE), and Symmetric Mean Absolute Percentage Error (SMAPE), where $RMSE=\sqrt{(1/H)\sum_{i=1}^H(y_i-\hat{y_i})^2}$, $MAE=(1/H)\sum_{i=1}^H\|y_i-\hat{y_i}\|$, and $SMAPE=(100\%/H)\sum_{i=1}^H(2\|y_i-\hat{y_i}\|/\|y_i+\hat{y_i}\|)$.
\emph{H} is the forecasting horizon and $H=48$ for the half-hourly data here.
$y_i$ and $\hat{y_i}$ are truth and predicted load at time \emph{i}.
These metrics are calculated for each time step of the test dataset, but they are then averaged over a whole test period and noted as $\overline{rmse}$, $\overline{mae}$, and $\overline{smape}$.

\subsection{Explanation of the model}
Previous research has confirmed that well-selected text features enhance forecasting, yet it is difficult to explain this improvement deeply.
Explainability is necessary to shed light on the behavior of the trained machine learning models, which would be completely black boxes otherwise. 
This paper attempts to explore the text-improving mechanisms from global and local correlations, and causality effects.
Global explainability is analyzed using Pearson correlation coefficients.  
Local explainability is analyzed using the Local Interpretable Model-agnostic Explanation (LIME), and causality is analyzed through Double Machine Learning (Double ML).

\subsubsection{Local explainability}
LIME targets a sample of the original data and generates a new, normally distributed local dataset using the current sample \cite{ribeiro2016should}.
In the next step, a simple surrogate model, such as linear regression, is used to fit the new dataset, yielding a locally interpretable perspective on the data under perturbation.
We can interpret how features in the current local affect the forecasting by viewing the coefficients in the linear model.

\subsubsection{Causality effects}
\label{Causation}
The Pearson correlation or the LIME model verifies the correlation between the candidate features and demand. 
Nevertheless, positive or negative correlations between variables may be the result of coincidence, and it is known that correlation does not imply causation \cite{aldrich1995correlations}.
For this reason, Double ML is proposed to verify the causality between the features and the target.
This is the effect of a particular feature of interest (Treatment, \emph{T}) on the predictions (Outcomes, \emph{Y}), provided that the rest of the features remain constant (Confounders, \emph{X}).
The causality is produced by the following partially linear model \cite{chernozhukov2018double}:

\begin{equation}
\label{formular double ml}
    Y-\hat{Y} = \tau(T-\hat{T}) + \delta \quad E(\delta|X,T)=0,
\end{equation}

in which $\hat{Y}=g(X)$ and $\hat{T}=h(X)$ are forecasts of \emph{Y} and \emph{T}, and $g(X)$ and $h(X)$ are nuisance functions that can be replaced by numerous machine learning methods.
In this case, we set $g(X)$ and $h(X)$ for all ExtraTrees regressors.
The treatment effect $\tau$ can be obtained with Ordinary Least Squares (OLS).

\subsection{Datasets}\label{Case Study}

In the field of forecasting with text, the first task is to choose two interconnected datasets: the forecasting target and an accompanying text corpus.
In our case, we chose the load and news datasets, considering that:
i) a geographical region is shared by the news audience and electricity consumers; 
ii) there is a considerable temporal overlap;
iii) the text is solid and authoritative (regarding the news);
iv) the language is accessible to authors and readers; 
v) open-access data is potentially comprised to facilitate replicability, and 
vi) electricity demand is utilized rather than net load data, which is essential to ensure coherent results and augment interpretability.
 
This work used four datasets covering five years, from June 2016 to May 2021. 
The first four years are used as a training set, and the last year is used as a test set.
Aggregated electric demands for the UK and Northern Ireland are obtained from the ENTSO-E transparency platform \cite{entso-e} along with the official day-ahead forecasts.
Historical bank holidays and daily temperatures for London were taken from commercial websites. 
According to \cite{obst2021adaptive}, we used the observed rather than predicted temperatures for the convenience of reproducibility.

Previous studies have mainly used keywords for external texts to filter news related to the forecasting domain \cite{li2019text,bai2022crude}.
However, this paper proposes to use the entire volume of news from the British Broadcasting Corporation (BBC) to explore the impact of broader social events on electricity load forecasting.
Over 80,000 news items were collected by the repository \cite{News-crawler} archived in \cite{BBC_news}.

\section{Results}\label{Results}
\subsection{Benchmark model}\label{Benchmark model}
A prediction model based on ExtraTrees Regression (ETR) was trained on the primary datasets.
Grid search and five-fold cross-validation were used to identify the optimal parameters and avoid overfitting.
The performance of the benchmark model is summarized in Table~\ref{Forecasting errors for models with H,W,T}, in which different combinations of features are tested: demand $\mathcal{D}$, calendar features $\mathcal{C}$ and temperature $\mathcal{T}$. 
These are compared with the official forecasts obtained from the ENTSO-E Transparency Platform.

\begin{table}[ht]
\centering
\caption{Benchmark model performance}
\label{Forecasting errors for models with H,W,T}
\renewcommand\arraystretch{1.3}
\begin{tabular}{p{1.9cm}ccc}
\hline
Features   & $\overline{rmse}$(MW) & $\overline{mae}$(MW) & $\overline{smape}$(\%)     \\ \hline
$\mathcal{ENTSO-E}$  & 2800.50     & 2544.86     & 7.65 \\
$\mathcal{D}$       & 2983.05±5.35 & 2539.25±5.14 & 7.75±0.01 \\
$\mathcal{D}+\mathcal{C}$     & 2896.49±4.81 & 2468.48±4.22 & 7.56±0.01 \\
$\mathcal{D}+\mathcal{T}$     & 2938.40±2.81 & 2488.93±2.61 & 7.59±0.01 \\
$\mathcal{D}+\mathcal{C}+\mathcal{T}$    & 2800.77±4.84 & \textbf{2374.07±4.39} & \textbf{7.29±0.01} \\

\hline
\end{tabular}
\end{table}

According to Table~\ref{Forecasting errors for models with H,W,T}, combining the electricity demand lags, calendars, and temperatures produces performance close to the official forecasts over the three deterministic metrics. 
However, holidays do not provide improvements. 
The best improvements are around 200MW in $\overline{mae}$ and 4\% in $\overline{smape}$.

Note that the target is not to develop forecasting models better than ENTSO-E models, which are mature and have long been used as a baseline in numerous studies.
This paper investigates the connections between news and load demand and explores how the potential text information improves forecasting.
It is also difficult to reproduce the ENTSO-E method, which is not publicly available, with limited information.
We therefore built an alternative model with a similar performance, by which we can compare the results with different textual features, and which also provides the possibility to improve the official forecasts.

\begin{figure*}[ht]
\centering
\includegraphics[scale=.12]{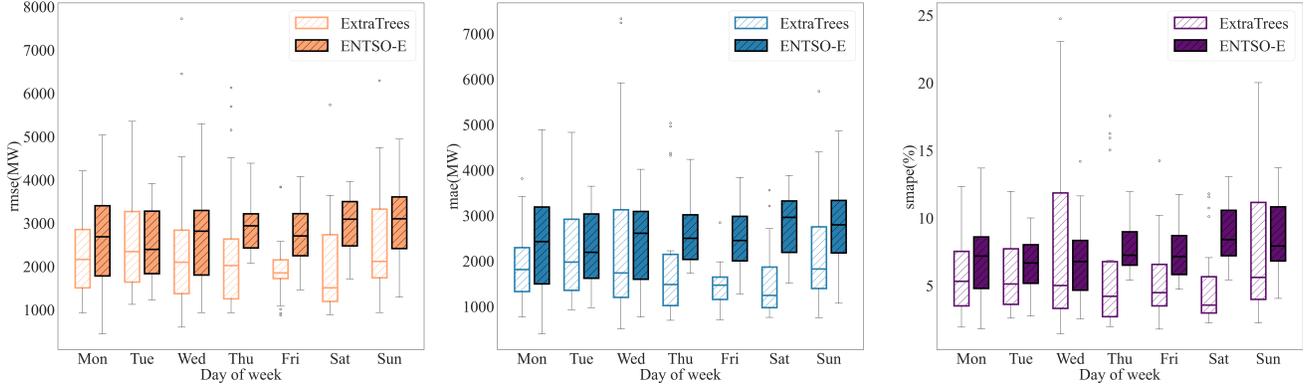}
\caption{Day of week averaged error box plots for the ExtraTrees and ENTSO-E method.
Left: averaged $\overline{rmse}$(MW); 
middle: averaged $\overline{mae}$(MW); 
right: averaged $\overline{smape}$(\%).
The labels on the x-axis are the days of the week, and those on the y-axis are metric values.}
\label{ExtraTree_ENTSO}
\end{figure*}

We averaged the forecasting metrics of the ExtraTrees model ($\mathcal{D}+\mathcal{C}+\mathcal{T}$ in Table~\ref{Forecasting errors for models with H,W,T}) and ENTSO-E method by day of week over the test period, and plotted the box plots in Figure~\ref{ExtraTree_ENTSO}.
In general, Figure~\ref{ExtraTree_ENTSO} presents the distributions across all $\overline{rmse}$, $\overline{mae}$, and $\overline{smape}$ values, including the medians and other percentiles, and the medians of the ExtraTrees model are lower than that of ENTSO-E.
Figure~\ref{ExtraTree_ENTSO} also supports that the ExtraTrees could be a promising alternative to the ENTSO-E method, aimed at facilitating more streamlined and efficient execution of the ablation studies.
We refer to the ExtraTrees as the benchmark in what follows.

\subsection{Textual feature-enhanced model}\label{Deterministic forecasting}
\subsubsection{Impact of textual features}
\label{Impact of textual features}

We trained a new model based on the benchmark ($\mathcal{D+C+T}$) with textual features $\mathcal{F}_t$ extracted from the news Titles (\emph{\textbf{T}}), Descriptions (\emph{\textbf{D}}) and Bodies (\emph{\textbf{B}}), including Count Features ($\mathcal{CF}$), Words Frequencies ($\mathcal{WF}$), Sentiment ($\mathcal{SE}$), Topic Distributions ($\mathcal{TD}$), and GloVe Word Embeddings ($\mathcal{GWE}$).
The experimental results are presented in Table~\ref{Forecasting errors when adding Ft into model D+C+T} with the original and selected feature numbers after the Granger test.
\added{For simplicity, we only list the $\mathcal{F}_t$ groups that improve forecasting.}
Table~\ref{Textual features description} lists the detailed features within each $\mathcal{F}_t$ group.

\begin{table}[ht]
\centering
\caption{Forecasting errors when adding $\mathcal{F}_t$ into benchmark model}
\label{Forecasting errors when adding Ft into model D+C+T}
\renewcommand\arraystretch{1.3}
\scalebox{0.92}{ 
\begin{threeparttable} 
\begin{tabular}{lcccc}
\hline
$\mathcal{F}_t$ Groups  & \#Selected $\mathcal{F}_t$ &  $\overline{rmse}$(MW) & $\overline{mae}$(MW) & $\overline{smape}$(\%)           \\ \hline
Benchmark                       & 0(0)                     & 2800.77±4.84 & 2374.07±4.39 & 7.29±0.01 \\ 
$\mathcal{WF_T}$   & 32(456)   & 2702.23±4.95 & 2283.52±4.24 & 6.98±0.01\\
$\mathcal{WF_D}$   & 25(329)  & 2760.65±6.85 & 2342.98±6.16 & 7.14±0.02 \\
$\mathcal{WF_B}$   & 10(550)  & 2751.40±4.38 & 2330.54±3.95 & 7.16±0.01 \\
$\mathcal{SE_B}$   & 1(18)    & 2788.45±4.61 & 2360.75±4.33 & 7.26±0.01 \\
$\mathcal{TD_B}$   & 2(69)    & 2747.51±5.60 & 2323.50±5.36 & 7.12±0.02 \\
$\mathcal{GWE_B}$  & 3(100)   & 2749.97±2.31 & 2327.66±2.43 & 7.16±0.01 \\ \hline
\end{tabular}
\end{threeparttable}  
 }
\begin{tablenotes}   
  \footnotesize            
  \item (i) \# is the counting numbers. 
  \item (ii) The subscripts inside $\mathcal{F}_t$ Groups are text types \emph{\textbf{T}}, \emph{\textbf{D}}, and \emph{\textbf{B}}. 
  \item (iii) In \#Selected $\mathcal{F}_t$, (\#) out of \# features are selected.
\end{tablenotes} 
\end{table}

\begin{table}[ht]
\centering
\caption{Textual features description}
\label{Textual features description}
\begin{tabular}{ll}
\hline
$\mathcal{F}_t$ Groups  & Features 
\\ \hline
$\mathcal{WF}_T$  &  \vtop{\hbox{\strut Andy, Murray, newspaper, headlines, rules, Ireland, year, }\hbox{\strut NHS, staff, Glasgow, family, home, European, mark, }\hbox{\strut updates, Paris, say, elections, premier, hit, bomb, second, }\hbox{\strut funeral, talks, Spain, budget, driver, care, sorry, Scotland,} \hbox{\strut job, coronavirus}}   \\ 
$\mathcal{WF}_D$  & \vtop{\hbox{\strut following, around, Ireland, MPs, least, away, reach, }\hbox{\strut schools, wife, shows, weeks, help, figures, days, lead,}\hbox{\strut Wales, security, hit, outside, Scotland, Monday, leaders,}\hbox{\strut restrictions, pandemic, coronavirus}} \\ 
$\mathcal{WF}_B$ &  \vtop{\hbox{\strut coming, power, city, inside, job, ahead, social, strong,}\hbox{\strut return, war}} 
\\ 
$\mathcal{SE}_B$ & Minimum subjectivity value     \\ 
$\mathcal{TD}_B$  & \vtop{\hbox{\strut \textbf{Topic-5 (Politics related to Northern Ireland)}: Ireland,}\hbox{\strut Northern, Irish, Belfast, DUP, Foster, republic, border, }\hbox{\strut Sinn, Neill}}  \\
& \vtop{\hbox{\strut \textbf{Topic-18 (Coronavirus)}: covid, coronavirus, pandemic,} \hbox{\strut cummings, Downing, street, question, adviser, Johnson,} \hbox{\strut questions}}\\ 
$\mathcal{GWE}_B$ & \vtop{\hbox{\strut \textbf{Dim-9 (Weapons)}: Hossein, warhead, Gangnam, }\hbox{\strut interceptor, missiles, bomb, enriched, clerical, Quds, }\hbox{Ballistic}}   \\
& \vtop{\hbox{\strut \textbf{Dim-51 (Transportation)}: Persia, Ibn, Arriva, }\hbox{\strut Transpennine, Merseyrail, fax, Mesopotamia, BBBofC, }\hbox{Crosscountry, Daren}}  \\
& \vtop{\hbox{\strut \textbf{Dim-69 (Military)}: ang, corps, Muhammadu,}\hbox{\strut commandant, army, commander, graduated, ante, military, }\hbox{\strut Buhari}} \\ \hline
\end{tabular}
\end{table}

The results in Table~\ref{Forecasting errors when adding Ft into model D+C+T} show that six groups of $\mathcal{F}_t$ improve the benchmark.
$\mathcal{WF}$ in \emph{\textbf{T}}, \emph{\textbf{D}}, and \emph{\textbf{B}} reduces forecasting errors, especially in \emph{\textbf{T}}.
In addition, $\mathcal{SE}$, $\mathcal{TD}$ and $\mathcal{GWE}$ from \emph{\textbf{B}} are better than those in \emph{\textbf{T}} and \emph{\textbf{D}}.
The features in Table~\ref{Textual features description} cover more information.
Regarding the $\mathcal{WF}$, it includes the name of a famous tennis player (Andy Murray), countries and cities in Europe (Ireland, Glasgow, Paris, Spain, Scotland, Wales), politics (elections, MPs-Members of Parliament, leaders), and COVID-19 (coronavirus, pandemic).
In $\mathcal{SE}$, the minimum subjectivity value is more influential and can be understood as the lower bound of the personal opinion contained in the daily news.
$\mathcal{TD}$ includes content about politics related to Northern Ireland and coronavirus.
$\mathcal{GWE}$ are about weapons, transportation, and military issues.
In conclusion, the impact of social aspects on load demand should be addressed, especially news concerning hot issues, political dynamics, global disasters, and geopolitical conflicts.

\subsubsection{Feature combination}
Different $\mathcal{F}_t$ group combinations are tested in this subsection to improve the performance further, including $\mathcal{M}_0 := \mathcal{WF}_T$,
$\mathcal{M}_1 := \mathcal{WF}_T+\mathcal{WF}_D+\mathcal{WF}_B$,
$\mathcal{M}_2 := \mathcal{WF}_T+\mathcal{SE}_B$,
$\mathcal{M}_3 := \mathcal{WF}_T+\mathcal{TD}_B$,
$\mathcal{M}_4 := \mathcal{WF}_T+\mathcal{GWE}_B$,
$\mathcal{M}_5 := \mathcal{WF}_T+\mathcal{SE}_B+\mathcal{TD}_B$,
$\mathcal{M}_6 := \mathcal{WF}_T+\mathcal{SE}_B+\mathcal{GWE}_B$,
$\mathcal{M}_7 := \mathcal{WF}_T+\mathcal{TD}_B+\mathcal{GWE}_B$,
$\mathcal{M}_8 :=\mathcal{WF}_T+\mathcal{SE}_B+\mathcal{TD}_B+\mathcal{GWE}_B$.

\begin{figure*}[ht]
\centering
\includegraphics[scale=.12]{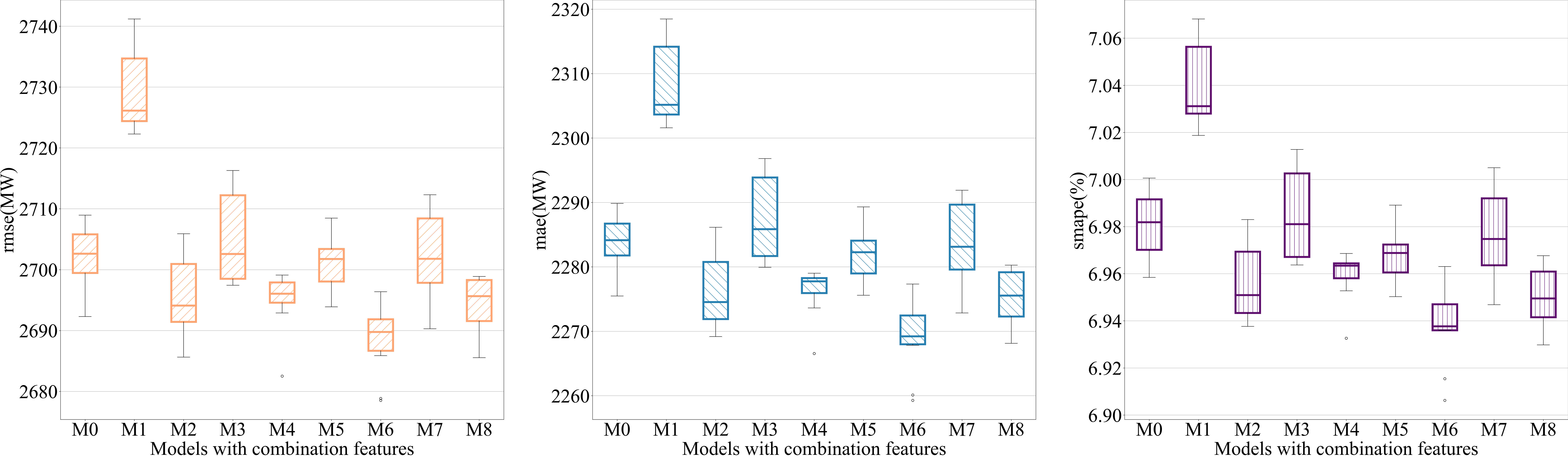}
\caption{Errors box plots for feature combination.
Left: errors for $\overline{rmse}$(MW); 
middle: errors for $\overline{mae}$(MW); right: errors for  $\overline{smape}$(\%).}
\label{Boxplot_combination}
\end{figure*}

Figure~\ref{Boxplot_combination} plots the results and illustrates that the forecasting errors increase when combining $\mathcal{WF}_D$ and $\mathcal{WF}_B$ with $\mathcal{WF}_T$, which demonstrates that $\mathcal{WF}_T$ is sufficient for forecasting in terms of $\mathcal{WF}$.
$\mathcal{M}_2$ shows the addition of $\mathcal{SE}_B$ can reduce forecasting errors.
$\mathcal{GWE}_B$ in $\mathcal{M}_4$ reduce the error spread in the box plot. 
$\mathcal{M}_6$, obtained from  $\mathcal{SE}_B$ and $\mathcal{GWE}_B$, combines the advantages of both and has the best performance, and is therefore used for further analysis.

\subsubsection{Error analysis and DM-test}
Errors are analyzed according to different hours by the model $\mathcal{M}_6$, and the results are shown in Figure~\ref{RMSE, MAE, and SMAPE on different hours}. 
The performance improvement is generally more remarkable in the early morning and evening, usually characterized by much larger ramps in demand.

\begin{figure}[ht]
  \centering
  \includegraphics[scale=0.09]{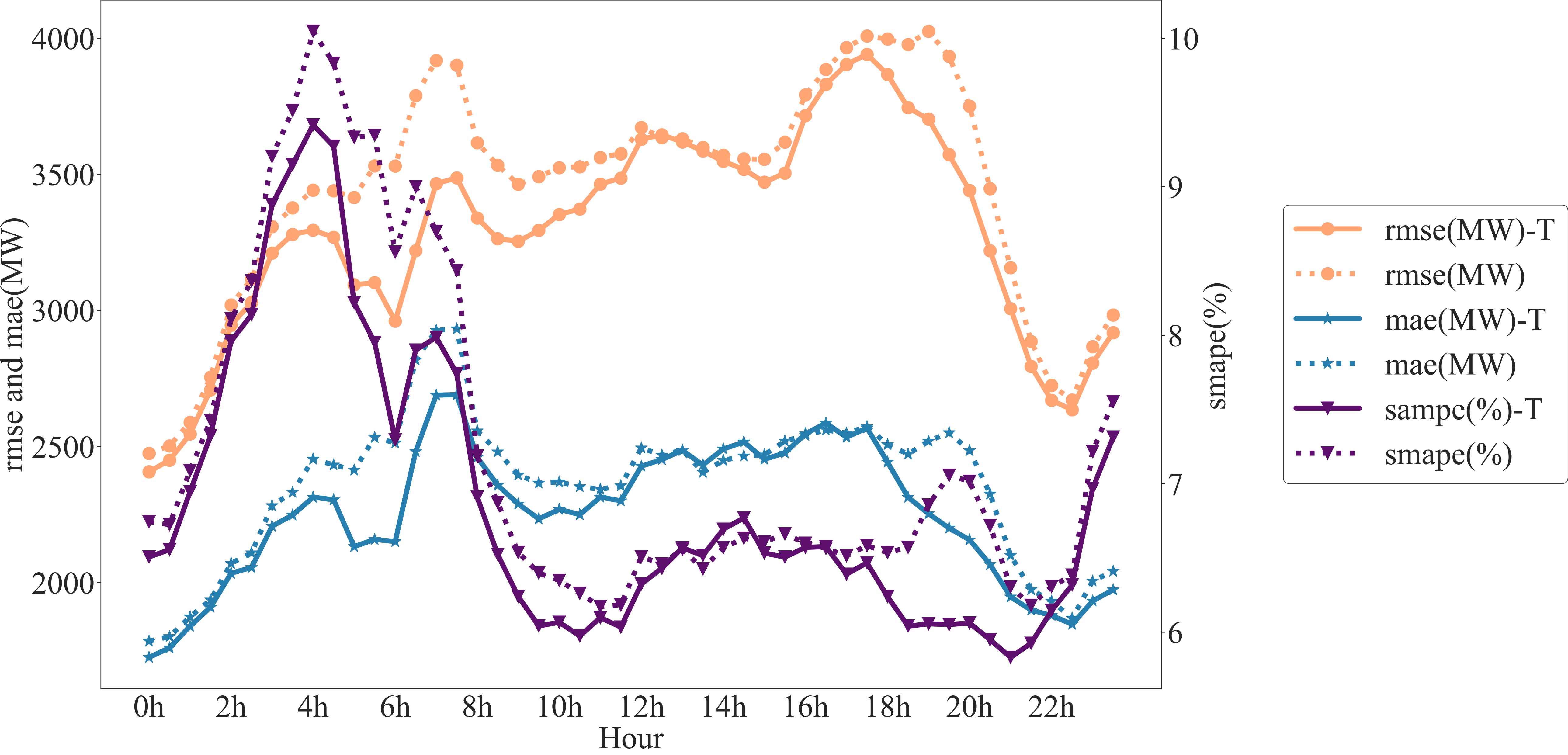}
  \caption{$\overline{rmse}$, $\overline{mae}$, and $\overline{smape}$ for different hours. The dashed and solid lines show forecasting without and with textual features.}
  \label{RMSE, MAE, and SMAPE on different hours}
\end{figure}

The Diebold-Mariano (DM) test is applied to the forecasts across models to statistically evaluate the model difference.
The null hypothesis $\mathcal{H}_0$ is that the two models are not significantly different. 
Given the one-sided situation, the alternative hypothesis is that one model is better than the other.
With a p-value of less than 0.05, we can infer a significantly better model.
We took four models for the DM-test: the ETR $\mathcal{D}$, benchmark $\mathcal{D+C+T}$ in Table~\ref{Forecasting errors for models with H,W,T}, $\mathcal{M}_0$ and $\mathcal{M}_6$ shown in Figure~\ref{Boxplot_combination}.

Table~\ref{P-values of DM-test results} shows the p-values of the DM-test for the four models.
The bold p-values are less than 0.05, for which we reject the null hypothesis and select the model in the column, which is better than the one in the row.
For example, when comparing the model $\mathcal{D+C+T}$ and $\mathcal{M}_0$, the p-value is 0.0404 and less than 0.05.
Therefore, we establish that there is a statistically significant difference between the forecasting accuracy of model $\mathcal{M}_0$ and $\mathcal{D+C+T}$, and that $\mathcal{M}_0$ is superior to $\mathcal{D+C+T}$.

\begin{table}[ht]
\centering
\caption{P-values of DM-test results}
\label{P-values of DM-test results}
\begin{tabular}{lllll}
\hline
      & $\mathcal{D}$       & $\mathcal{D+C+T}$               & $\mathcal{M}_0$             & $\mathcal{M}_6$              \\ \hline
$\mathcal{D}$      & 1.0000      & \textbf{0.0000**} & \textbf{0.0000**} & \textbf{0.0000**} \\
$\mathcal{D+C+T}$   & 0.9999 & 1.0000                 & \textbf{0.0404*}  & \textbf{0.0181*}  \\
$\mathcal{M}_0$ & 0.9999 & 0.9596            & 1.0000                 & \textbf{0.0322*}  \\
$\mathcal{M}_6$  & 0.9999 & 0.9818            & 0.9678            & 1.0000                 \\ \hline
\end{tabular}
\begin{tablenotes}   
  \footnotesize            
  \item[1] * for $0.01<p<0.05$, and ** for $0<p \leq 0.01$
\end{tablenotes} 
\end{table}

\subsubsection{Results on short-term forecasting}

In practice, short-term demand forecasting (a few hours ahead), is important in electricity grid operations, market trading strategies, and energy storage system management \cite{yang2019short}. 
This subsection evaluates and compares the short-term forecasting of the ENTSO-E method, benchmark ETR, and the model $\mathcal{M}_6$.
We considered the first two hours of the day-ahead forecasting: midnight and 1 a.m.
In the next step, the error metrics were calculated for those two hours each day and averaged over the whole test set, as shown in Table~\ref{Short-term forecasting results comparison}.

\begin{table}[ht]
\centering
\caption{Short-term forecasting results comparison}
\label{Short-term forecasting results comparison}
\begin{tabular}{p{1.9cm}ccc}
\hline
Models     & $\overline{rmse}$(MW) & $\overline{mae}$(MW) & $\overline{smape}$(\%) \\ \hline
ENTSO-E   & 2484.74  & 2423.56 & 9.27      \\
ETR & 1892.49  & 1844.55 & 6.99      \\
$\mathcal{M}_6$        & 1841.04  & 1793.21 & 6.78      \\ \hline
\end{tabular}
\end{table}

Table~\ref{Short-term forecasting results comparison} illustrates that the short-term forecasting errors from all the mentioned methods are lower than those of day-ahead forecasting.
$\mathcal{M}_6$ improves ETR by around 3\%, showing that the news textual features from the previous day can enhance short-term day-ahead forecasting.

\subsection{Textual feature variation and sensitivity analysis}
\label{Textual Feature Variation and Sensitivity Analysis}
This subsection provides a sensitivity analysis for the process of extracting textual features from the news, considering the news amount and word embedding methods.

\subsubsection{Variation in the amount of news}
Previous experiments considered all of the BBC news available, without a pre-selection of contents.
For example in \cite{bai2022crude}, only news related to specific futures was used in oil/gas/gold price forecasting.
The selected text often relates to the forecasting target, and the training time is reduced due to a smaller amount of text.
Thus, the trade-off between retaining more information and reducing training time is worth studying.
In the following, we show how news volumes impact forecast precision, and how the news bodies are selected to simplify the process.

Besides the news text, the BBC website also provides the sections of the current news.
We selected the top 10 sections with the most news, then searched and saved news with the root `\emph{electric-}', and created an electricity-related section.
All of the information can be found in Table~\ref{Statistic information of the section news}.

\begin{table}[ht]
\centering
\caption{Statistic information of the section news}
\label{Statistic information of the section news}
\begin{tabular}{lcccc}
\hline
Section Name          & \# news & \# days & \% news & \% days \\ \hline
UK                    & 8386    & 1674    & 10.40                & 91.68                \\
Business              & 6575    & 1517    & 8.15                 & 83.08                \\
UK Politics           & 5703    & 1532    & 7.07                 & 83.90                \\
Entertainment \& Arts & 3950    & 1538    & 4.90                 & 84.23                \\
US \& Canada          & 3633    & 1442    & 4.50                 & 78.97                \\
Europe                & 2860    & 1359    & 3.55                 & 74.42                \\
London                & 2456    & 1217    & 3.05                 & 66.65                \\
Wales                 & 2308    & 954     & 2.86                 & 52.25                \\
Health                & 2205    & 1161    & 2.73                 & 63.58                \\
Northern Ireland      & 1410    & 791     & 1.75                 & 43.32                \\
Electric              & 1771    & 1077    & 2.20                 & 58.98                \\ \hline
\end{tabular}
\begin{tablenotes}   
  \footnotesize            
  \item (i) \# for the number of news or days.
  \item (ii) \% for the percentage of the section news count and days to the \\total counts (80650) and days (1826).
\end{tablenotes} 
\end{table}

All of the $\mathcal{F}_t$ groups were aggregated by the same section, and then used in the demand forecasting task.
Table~\ref{Forecasting errors with the Ft groups from the section news} illustrates the results for the $\mathcal{F}_t$ groups and compares the benchmark, the model with text from all the sections, and models with text from a single section.
For simplicity, only models that perform better than the benchmark are shown in Table~\ref{Forecasting errors with the Ft groups from the section news}.
The bold results are the best within a certain $\mathcal{F}_t$ group.

\begin{table}[ht]
\centering
\caption{Forecasting errors with the $\mathcal{F}_t$ groups from the sectional news}
\label{Forecasting errors with the Ft groups from the section news}
\begin{tabular}{llccc}
\hline
\multicolumn{2}{l}{Sections}         & $\overline{rmse}$(MW)                  & $\overline{mae}$(MW)                   & $\overline{smape}$(\%)              \\ \hline
\multicolumn{2}{l}{Benchmark}        & 2800.77±4.84          & 2374.07±4.39          & 7.29±0.01          \\ \hline
\multirow{4}{*}{$\mathcal{CF}$}  & All sections  & 2798.81±4.85          & 2371.95±4.51          & 7.29±0.01          \\
                     & Europe        & \textbf{2783.59±6.27} & 2358.79±5.65          & 7.25±0.02          \\
                     & Wales         & 2785.54±4.58          & \textbf{2355.72±3.78} & \textbf{7.24±0.01} \\
                     & Electricity   & 2784.48±7.48          & 2360.13±6.95          & 7.25±0.02          \\ \hline
\multirow{8}{*}{$\mathcal{WF}$}  & All sections  & \textbf{2751.40±4.38} & \textbf{2330.54±3.95} & \textbf{7.16±0.01} \\
                     & UK Politics   & 2795.15±5.47          & 2367.73±5.59          & 7.28±0.02          \\
                     & Europe        & 2794.51±4.42          & 2367.70±4.41          & 7.25±0.01          \\
                     & Entertainment & 2794.44±4.56          & 2371.37±4.11          & 7.29±0.01          \\
                     & US\&Canada    & 2779.29±4.20          & 2354.78±4.01          & 7.24±0.01          \\
                     & Wales         & 2769.10±8.14          & 2346.03±7.26          & 7.17±0.02          \\
                     & London        & 2781.53±4.34          & 2361.34±3.96          & 7.26±0.01          \\
                     & NIE           & 2772.37±6.14          & 2349.02±5.41          & 7.22±0.02          \\ \hline
$\mathcal{SE}$                  & All sections  & \textbf{2788.45±4.61} & \textbf{2360.75±4.33} & \textbf{7.26±0.01} \\ \hline
\multirow{11}{*}{$\mathcal{TD}$} & All sections  & 2747.51±5.60          & 2323.50±5.36          & 7.12±0.02          \\
                     & UK            & 2780.51±3.55          & 2357.19±3.02          & 7.24±0.01          \\
                     & UK Politics   & 2770.32±5.34          & 2346.37±5.22          & 7.21±0.01          \\
                     & Europe        & 2790.55±5.41          & 2367.26±5.24          & 7.26±0.02          \\
                     & Entertainment & 2781.41±6.23          & 2358.11±6.05          & 7.25±0.02          \\
                     & US\&Canada    & 2764.20±3.82          & 2342.82±3.72          & 7.20±0.01          \\
                     & Wales         & 2789.82±2.92          & 2364.09±2.80          & 7.27±0.01          \\
                     & London        & 2782.26±4.24          & 2358.55±4.39          & 7.24±0.01          \\
                     & Health        & 2792.97±4.48          & 2365.44±4.37          & 7.28±0.01          \\
                     & NIE           & \textbf{2717.55±4.87} & \textbf{2304.13±4.26} & \textbf{7.07±0.01} \\
                     & Electricity   & 2785.59±6.98          & 2360.42±6.23          & 7.26±0.02          \\ \hline
\multirow{4}{*}{$\mathcal{GWE}$} & All sections  & \textbf{2749.97±2.31} & 2327.66±2.43          & \textbf{7.16±0.01} \\
                     & Wales         & 2766.41±9.72          & 2339.72±9.11          & 8.03±0.03          \\
                     & London        & 2748.29±5.26          & \textbf{2320.73±4.18} & \textbf{7.97±0.01} \\
                     & Electricity   & 2771.60±10.11         & 2345.11±8.96          & 8.05±0.03          \\ \hline
\end{tabular}
\end{table}

Table~\ref{Forecasting errors with the Ft groups from the section news}, $\mathcal{CF}$ shows the sections \emph{Europe}, \emph{Wales}, and \emph{Electricity} that pass the Granger test and improve the performance against the benchmark, with a marginal improvement less than 1\%.
$\mathcal{WF}$ from seven sections improve the benchmark, although it is still preferable to consider all the sections.
Although $\mathcal{SE}$ from most of the sections pass the Granger test, none of them can exceed the results from all of the sections.
$\mathcal{TD}$ from every section except \emph{Business} perform better than the benchmark.
The topics from the Northern Ireland (NIE) section perform better than the results from the news as a whole.
They can be summarized as \textbf{Financial fraud} (church, money, Saudi, bank, fraud, data, security, cash, account, cyber), \textbf{Education} (students, university, education, school, college, results, exams, grades, teachers, pupils), \textbf{Government response to Covid-19} (Covid, coronavirus, pandemic, Cummings, Downing, Street, question, adviser, Johnson, Hancock), \textbf{Vaccine} (vaccine, cases, deaths, virus, health, England, hospital, vaccination, figures, care), and \textbf{Immigration in Europe} (French, France, president, migrants, immigration, countries, Europe, border, European, Paris).
For $\mathcal{GWE}$, although sections \emph{Wales}, \emph{London}, and \emph{Electricity} can improve the benchmark in $\overline{rmse}$ and $\overline{mae}$, they have a higher standard deviation than the results from all the news.

In conclusion, the information embedded in all the news can enhance forecasting, but news selected based on prior knowledge may not be helpful.
Using all the available news brings more knowledge, is essential for discovering broader social aspects impacting load, and provides more accurate results.
Despite the time spent training NLP models for all the news, on the other side, projecting its use in an eventual application is easily feasible with the current computing infrastructure.
The number of news items per day amounts to tens or hundreds, and the textual features can be calculated based on the pre-trained models.

\subsubsection{Comparing different word-embedding methods}
This subsection compares GloVe word embedding with two popular embedding methods: Word2Vec and Term Frequency-Inverse Document Frequency (TF-IDF).
Word2Vec can project the words in a large corpus into a continuous vector space by predicting the target word from the surrounding words or predicting the surrounding words from the target word in a context window \cite{mikolov2013efficient}.
TF-IDF is a measurement used to quantify the importance of the words in documents and identify rare words but with higher frequency \cite{sparck1972statistical}.
Term Frequency (TF) gives higher weights to words that appear frequently in documents, and Inverse Document Frequency (IDF) gives higher weights to rare words that appear in specific documents. 
In general, TF-IDF only captures the importance of words in the document without semantic relationships.
Word2Vec considers semantic relationships but is limited in the local view. 
Besides, GloVe incorporates global semantic relationships when training the model and provides pre-trained word vectors that can be easily used in different fields.

This paper utilized \proglang{TfidfVectorizer} in \proglang{sklearn 1.0.2} for the TF-IDF vectorizing, and \proglang{Word2Vec} in \proglang{gensim 4.1.2} for the Word2Vec vectorizing.
In Word2Vec, we set the dimension of the word vectors at 100, in line with that of GloVe.
In TF-IDF, the threshold is set to 5e-4, which saves those words with a TF-IDF value higher than the threshold, and 91 features are retained.
Table ~\ref{Forecasting errors of different embeddings} presents the forecasting results of the benchmark and GloVe from Table~\ref{Forecasting errors when adding Ft into model D+C+T}, adding the Word2Vec and TF-IDF on the news bodies.

\begin{table}[ht]
\centering
\caption{Forecasting errors of different embeddings}
\label{Forecasting errors of different embeddings}
\renewcommand\arraystretch{1.3}
\scalebox{0.92}{
\begin{tabular}{lccccc}
\hline
Embeddings  & \#Selected $\mathcal{F}_t$ &  $\overline{rmse}$(MW) & $\overline{mae}$(MW) & $\overline{smape}$(\%)           \\ \hline
Benchmark & 0(0) &2800.77±4.84 & 2374.07±4.39 &  7.29±0.01 \\
GloVe & 3(100) & 
\textbf{2749.97±2.31} & \textbf{2327.66±2.43} & 7.16±0.01 \\
Word2Vec & 6(100) & 2774.48±6.48 & 2350.42±6.08 & 7.22±0.02 \\ 
TF-IDF & 5(91) & 2752.49±5.96 & 2336.62±5.79 &  \textbf{7.13±0.02} \\ \hline

\end{tabular}
}
\end{table}

In Table~\ref{Forecasting errors of different embeddings}, adding information from each of the three embeddings improves the benchmark model, with the best-performing one being the GloVe on the $\overline{rmse}$ and $\overline{mae}$, and TF-IDF embedding outperforming the GloVe on $\overline{smape}$.

\subsection{Application to NIE demand}
This subsection explores the case of NIE and provides a regional view.
The NIE demand dataset was collected in the ENTSO-E Transparency Platform, and the text dataset was still BBC news.
We used ETR as the main model frame and evaluated the performance by adding different groups of textual features.
The model with demand lags, calendar features, and temperatures from Belfast, the capital and largest city of NIE, served as the benchmark for this case.

The marks of the three text types (\emph{\textbf{T}}, \emph{\textbf{D}}, \emph{\textbf{B}}) and five $\mathcal{F}_t$ groups in Section~\ref{Impact of textual features} are kept here.
All of the text features have been filtered by the Granger test again, with the target variable replaced by NIE demand.
The $\mathcal{F}_t$ groups were trained separately, and the results are illustrated in Table~\ref{Forecasting errors NIE}.
To simplify, Table~\ref{Forecasting errors NIE} only contains those results that are better than the benchmark and best in \emph{\textbf{T}}, \emph{\textbf{D}}, and \emph{\textbf{B}}.

\begin{table}[ht]
\centering
\caption{Forecasting errors when adding $\mathcal{F}_t$ into benchmark model for NIE}
\label{Forecasting errors NIE}
\renewcommand\arraystretch{1.3}
\scalebox{0.92}{ 
\begin{threeparttable} 
\begin{tabular}{lcccc}
\hline
$\mathcal{F}_t$ Groups & \#Selected $\mathcal{F}_t$ & 
$\overline{rmse}$(MW) & $\overline{mae}$(MW) & $\overline{smape}$(\%)\\ \hline
Benchmark  & 0(0) & 85.86±0.19 & 70.52±0.17 & 8.15±0.02 \\
$\mathcal{CF_D}$  & 2(27) & 78.95±0.14 & 64.71±0.12 & 7.51±0.02 \\
$\mathcal{WF_T}$  & 23(456) & 79.32±0.16 & 64.93±0.12 & 7.52±0.01 \\
$\mathcal{SE_B}$  & 1(18)  & \textbf{78.84±0.16} & \textbf{64.64±0.16} & \textbf{7.49±0.02} \\ 
$\mathcal{TD_D}$  & 5(100) & 79.00±0.16 & 64.73±0.16 & 7.50±0.02 \\
$\mathcal{GWD_D}$ & 3(100) & 78.91±0.18 & 64.72±0.16 & 7.51±0.02 \\ \hline
\end{tabular}
\end{threeparttable}  
 }
\end{table}

The results in Table~\ref{Forecasting errors NIE} demonstrate that the selected textual features enhance demand forecasting in NIE.
This fact provides insights implying that national-level news still benefits regional demand forecasting.
In detail, the $\mathcal{WF}_T$ and the $\mathcal{SE}_B$ improve forecasting both on national and regional levels.
The $\mathcal{CF}_D$, $\mathcal{TD}_D$, and $\mathcal{GWE}_D$ are also helpful for forecasting.
All of the textual feature groups listed in Table~\ref{Forecasting errors NIE} improve the benchmark by around 8\% for all three metrics, while the percentage in the national case is only around 2\%.

The detailed features in each $\mathcal{F}_t$ group are presented in Table~\ref{Textual features description NIE}.
In $\mathcal{CF}$, the number of news items from sections \emph{Science\&Environment} and \emph{Wales} is related to the load demand.
$\mathcal{WF}$ covers social events (fight, street, inquiry, delay), news and media (updates, list), locations or football clubs (Spain, Liverpool), and COVID-19 (pandemic, cases).
In $\mathcal{SE}$, the standard deviation of polarity in the news body is influential, and this feature measures the dispersion of the sentiment polarity of all the news in a day.
$\mathcal{TD}$ are military celebrations, city development, crime and law enforcement, international politics, and social environment according to the top words.
In $\mathcal{GWE}$, the words with higher values are listed to represent the latent meaning of the dimension.
The selected dimensions are related to international conflict, Middle Eastern culture, and film arts.

\begin{table}[ht]
\centering
\caption{Textual features description for Northern Ireland}
\label{Textual features description NIE}
\begin{tabular}{ll}
\hline
$\mathcal{F}_t$ Group  & Features 
\\ \hline
$\mathcal{CF}_D$  & section\_Science\&Environment, section\_Wales \\ 
$\mathcal{WF}_T$  &  \vtop{\hbox{\strut faces, list, fight, updates, level, street, captain, old,}\hbox{George, car, shows, service, Spain, Liverpool, inquiry, }\hbox{\strut delay, race, search, general, weather, cases, spy, }\hbox{\strut pandemic}}   \\ 
$\mathcal{SE}_B$ & Standard deviation of polarity in news body     \\
$\mathcal{TD}_D$  & \vtop{\hbox{\strut \textbf{Topic-6 (Military celebrations)}: stage, tour, jab, army, }\hbox{\strut founder, squad, base, grace, Easter, oxygen}}  \\
& \vtop{\hbox{\strut \textbf{Topic-49 (City development)}: well, town, airport, }\hbox{\strut became, huge, Indian, rapper, Dominic, aircraft, process}} \\ 
& \vtop{\hbox{\strut \textbf{Topic-54 (Crime)}: man, end, accused, arrested, murder,}\hbox{\strut police, charged, killing, January, suspicion}} \\ 
& \vtop{\hbox{\strut \textbf{Topic-56 (International politics)}: faces, countries,} \hbox{\strut summit, shared, rest, drop, kill, Brazil, Florida, gay}} \\ 
& \vtop{\hbox{\strut \textbf{Topic-69 (Social environment)}: prison, parts, cost, list, }\hbox{\strut clear, rates, looking, might, soon, interest}} \\ 
$\mathcal{GWE}_D$ & \vtop{\hbox{\strut \textbf{Dim-12 (International conflict)}: amputated, bodies, truce,}\hbox{\strut OFSTED, abducted, Kashmir, Muslim, torn, cease,}\hbox{\strut ceasefire}}   \\
& \vtop{\hbox{\strut \textbf{Dim-52 (Middle Eastern culture)}: Netanyahu, Arabic, }\hbox{\strut Sheikh, Sharon, poem, shah, Egyptian, Begum, dollars, }\hbox{\strut billion}}  \\
& \vtop{\hbox{\strut\textbf{Dim-61 (Film arts)}: films, movies, memories,}\hbox{\strut photographs, pieces, film, stories, actors, tropical, artists}} \\ \hline
\end{tabular}
\end{table}

\section{Discussion}\label{Discussion}
The following subsections explain the relationships between textual features and electricity load, from global, local, and causality perspectives.

\subsection{Global correlations}
\label{Global correlations between textual features and electricity load}

\begin{figure*}[htp]
\centering
\includegraphics[scale=.054]{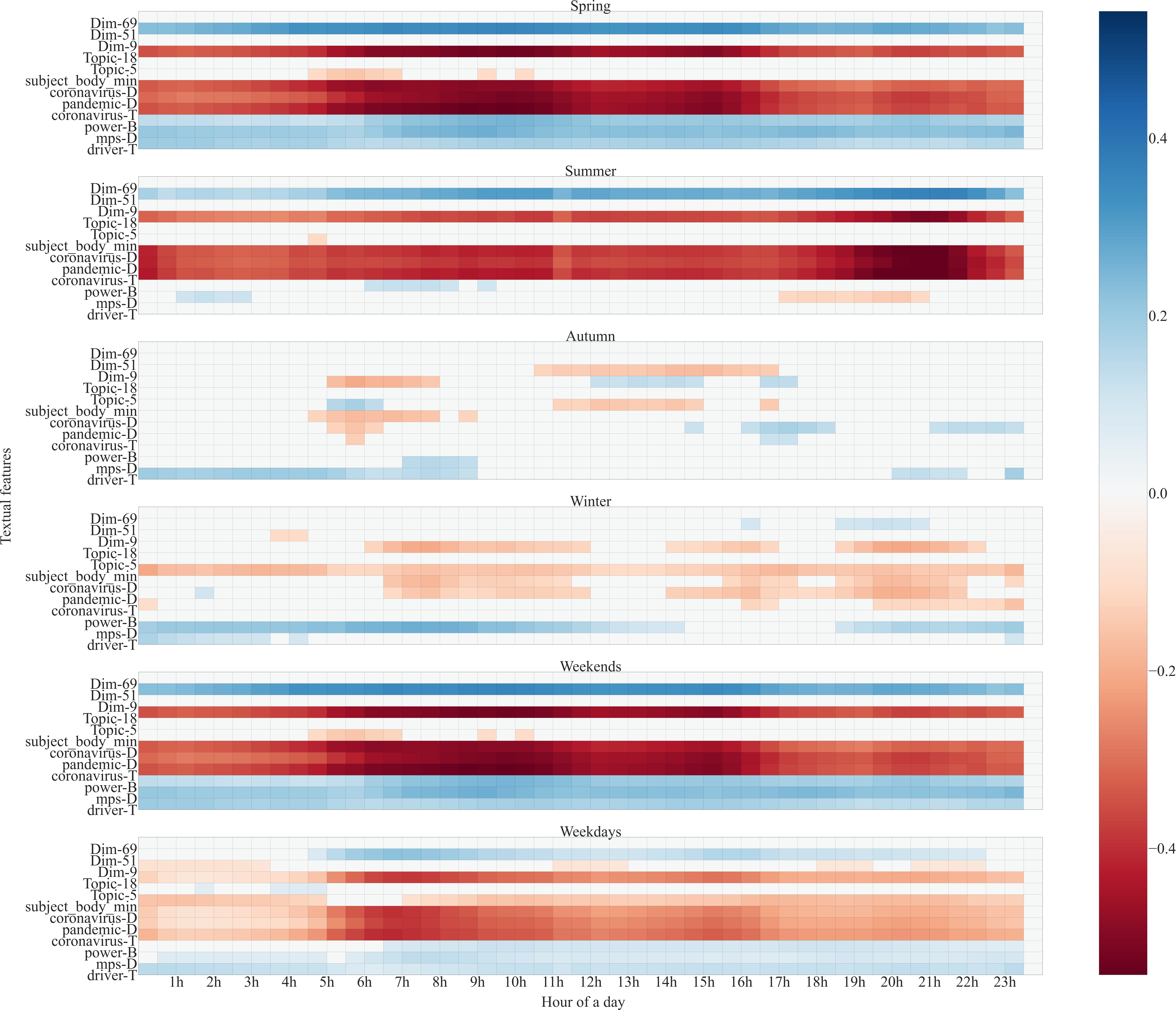}
\caption{Pearson coefficients for textual features and load. The coloured grids with $p<0.05$ stand for significant Pearson correlations, the blue ones correlate positively with the feature and the load at given hour, and the red ones stand for the negative correlations.}
\label{Pearson_analysis_NLP}
\end{figure*}

Table~\ref{Textual features description} introduced all the textual features that benefit forecasting. 
Due to a large number of words in $\mathcal{WF}_T$, we only mention six of the top 3 with the strongest positive and negative correlations overall.
In addition, $\mathcal{SE}_B$, $\mathcal{TD}_B$, and $\mathcal{GWE}_B$ were included.
Figure~\ref{Pearson_analysis_NLP} illustrates the correlations over hours on different seasons, weekdays, and weekends.

Generally, Figure~\ref{Pearson_analysis_NLP} presents more obvious correlations in spring, summer, and weekends.
Except for seasons correlations, \emph{driver-T} shows a positive correlation with load at dawn and in the early morning in autumn and winter.
The correlation is more pronounced for \emph{mps-D} in spring and winter than in other seasons.
The three words with negative correlations are all coronavirus-related. 
They show strong negative correlations with load in the spring and summer, with decreasing correlations in the subsequent seasons.
The coronavirus-related Topic-18 shows similar regularity.
We also noticed that the peaks differ in spring (daytime) and summer (evening), where the grids are darker.
In addition, social sentiment (\emph{subject\_body\_min} had a higher correlation in winter. 
Among the $\mathcal{GWE}$, only \emph{Dim-51}, train transportation in the UK, presents a positive correlation with the load in spring and summer and is more noticeable on weekends.
This correlation is relatively strong in the daytime in spring, and there is a peak from 19:00 to 22:00 in summer.

\subsection{Local correlations}
\label{Local correlations between textual features and electricity load}
\begin{figure}[ht]
\centering
\includegraphics[scale=.07]{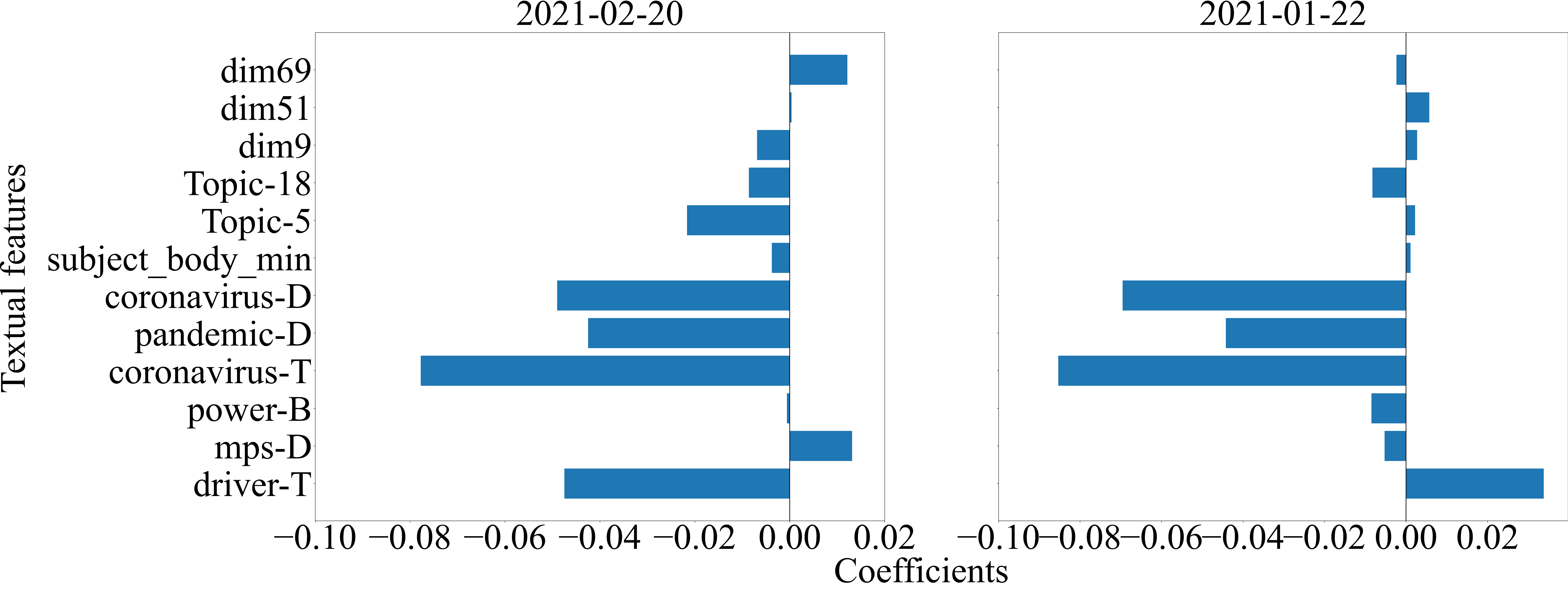}
\caption{Feature coefficients from LIME model}
\label{Local_LIME}
\end{figure}

Figure~\ref{Local_LIME} lists two days with the most negative (2021-02-20) and positive (2021-01-22) coefficients of \emph{driver-T}, for example.
Each subplot contains the textual features (y-axis) and their coefficients (x-axis) from the LIME model.
The word frequencies of \emph{coronavirus-T},  \emph{coronavirus-D}, and \emph{pandemic-D} serve negative roles on both days. 
At the same time, the coefficients of the other features vary, which is reasonable from a local view.

\subsection{Causality effects}
\label{Causality effects}
\subsubsection{Day-ahead forecasting}
This part tests the causality effects of all the textual features listed in Table~\ref{Textual features description}.
As described in Section~\ref{Causation}, $\tau>0$ indicates that the textual feature will positively cause a change of load demand, and vice versa.
We calculated all the $\tau$s for each textual feature on the 48 half-hours and retained those with $p<0.05$, and $\tau>0.10$, which are significant according to \cite{chernozhukov2021causal}.
Table~\ref{Words with significantly causality effects on load demand} illustrates all the load-positive and load-negative words from \emph{\textbf{T}}, \emph{\textbf{D}} and \emph{\textbf{B}} at different periods.

\begin{table}[ht]
\centering
\caption{Words with significant causality effects on load demand}
\label{Words with significantly causality effects on load demand}
\begin{tabular}{lll}
\hline
Time periods & Load-positive words                                                                                              & Load-negative words                                                                                            \\ \hline
0h-5h30      & \vtop{\hbox{\strut Andy-T, bomb-T,}\hbox{\strut days-D, elections-T,}\hbox{\strut pandemic-D, Paris-T,}\hbox{\strut year-T}}  & \vtop{\hbox{\strut ahead-B, hit-D, }\hbox{\strut power-B, second-T,}\hbox{\strut social-B}} \\ 

6h-11h30     & \vtop{\hbox{\strut Andy-T, bomb-T,}\hbox{\strut driver-T, elections-T,}\hbox{\strut headlines-T, Ireland-T,}\hbox{\strut lead-D, least-D, }\hbox{\strut outside-D, Paris-T,}\hbox{reach-D, sorry-T}}& 
\vtop{\hbox{\strut ahead-B, coming-B,}\hbox{\strut days-D, European-T,}\hbox{\strut figures-D, help-D,}\hbox{\strut job-B, pandemic-D,}\hbox{\strut return-B, social-B,}\hbox{\strut strong-B, Wales-D,}\hbox{\strut year-T}} \\ 

12h-17h30    & \vtop{\hbox{\strut coronavirus-D, hit-D,}\hbox{\strut hit-T, lead-D,}\hbox{\strut leaders-D, least-D,}\hbox{\strut premier-T}}& 
\vtop{\hbox{\strut ahead-B, coronavirus-T,}\hbox{\strut days-D, European-T,}\hbox{\strut figures-D, pandemic-D,}\hbox{\strut return-B, social-B,}\hbox{\strut Wales-D, year-T}}\\

18h-23h30    & 
\vtop{\hbox{\strut Andy-T,driver-T,}\hbox{\strut elections-T,family-T,}\hbox{\strut headlines-T, hit-T,}\hbox{\strut lead-D,newspaper-T,}\hbox{\strut outside-D,Paris-T,}\hbox{Scotland-D,sorry-T}}& 
\vtop{\hbox{\strut ahead-B,coming-B,}\hbox{\strut figures-D,pandemic-D,}\hbox{\strut power-B,Scotland-T,}\hbox{\strut social-B,Wales-D}}  \\ \hline
\end{tabular}
\end{table}

Table~\ref{Words with significantly causality effects on load demand} includes words with significant causality effects in four periods.
Similar to sentiment analysis, the words are categorized into load-positive and load-negative to reflect how the load responds to the changing word frequencies.
The load-positive words include famous tennis player, elections, and the city names.
The negative aspects are power and the pandemic.
The causality effects of the words are time-varying, and the same word can have opposite effects.
For example, ``pandemic" in the news description has a positive effect on demand at midnight, but negatively impacts demand during the rest of the day.

\begin{figure}[ht]
\centering
\includegraphics[scale=.18]{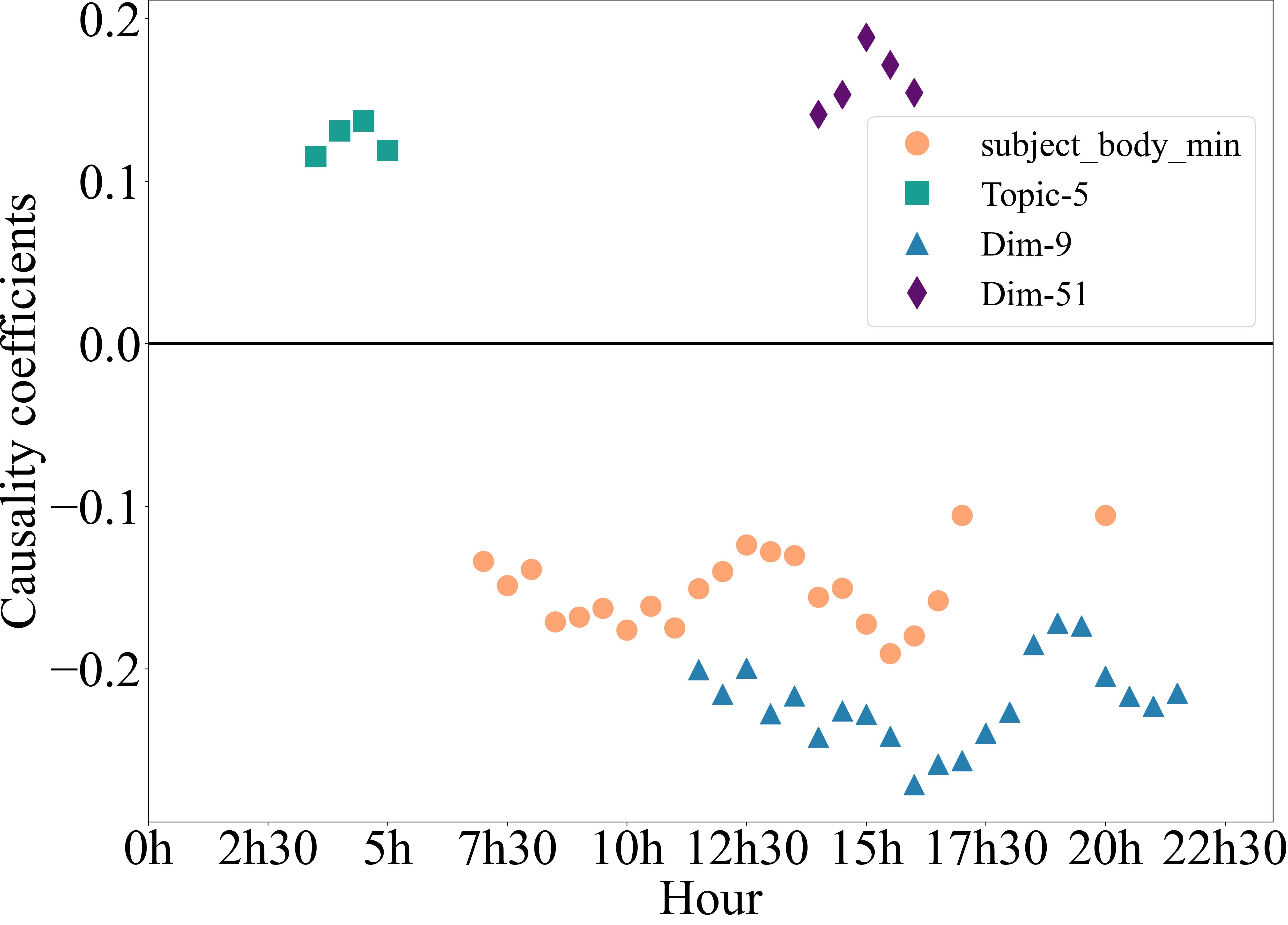}
\caption{Scatter plot for causality coefficients of the textual features}
\label{Scatter plot for causality}
\end{figure}

We then plotted all the $\tau$s for \emph{subject\_body\_min}, \emph{Topic-5}, \emph{Topic-18}, \emph{Dim-9}, \emph{Dim-51}, and \emph{Dim-69} at the 48 half-hours as in Figure~\ref{Scatter plot for causality}, which illustrates when and how they impact load demand.
\emph{Topic-5} (politics related to Northern Ireland) and \emph{Dim-51} (Transportation) have positive causal effects on demand in the early morning and afternoon, both in the very short term.
The \emph{subject\_body\_min}, also the sentiment feature, negatively impacts demand from morning to evening, with a causality effect around 15\% and aligned with human activity hours.
\emph{Dim-9} (weapon) has the causality effect higher than 20\% from noon to late night.

\subsubsection{Short-term forecasting}
We then completed a more granular analysis of the short-term forecasting.
We retrained the causality model only at midnight and 1 a.m. for the forecasting day, and kept the $\tau$s with $p<0.05$.
As the results only concern two hours (four half-hours), we only include these in Table~\ref{P-values and treatment affects of causality for short-term forecasting}.

\begin{table}[ht]
\centering
\caption{P-values and causality affects $\tau$s for short-term forecasting}
\label{P-values and treatment affects of causality for short-term forecasting}
\begin{tabular}{lcccccc}
\hline
\multirow{2}{*}{Time} & \multicolumn{2}{c}{power-B} & \multicolumn{2}{c}{pandemic-D} & \multicolumn{2}{c}{Topic-5} \\ 
                      & p-value       & $\tau$        & p-value              & $\tau$    & p-value       & $\tau$        \\ \hline
0h                    & 0.0039**      & -0.1335     & \textgreater{}0.05   & 0       & 0.0016**      & 0.1005      \\
0h30                  & 0.0122*       & -0.1076     & \textgreater{}0.05   & 0       & 0.0026**      & 0.0876      \\
1h                    & 0.0162*       & -0.1004     & 0.0316*              & 0.1296  & 0.0056**      & 0.0793      \\
1h30                  & 0.0243*       & -0.1063     & 0.0134*              & 0.1662  & 0.0082**      & 0.0839\\
\hline
\end{tabular}
\begin{tablenotes}   
  \footnotesize            
  \item * for $0.01<p<0.05$, and ** for $0<p \leq 0.01$.
\end{tablenotes} 
\end{table}

Only \emph{power-B}, \emph{pandemic-D}, and \emph{Topic-5} passed the causality test for the short-term forecasting.
\emph{Power-B} is related to electricity energy, power, or capacity.
It is reasonable that these features would negatively impact short-term load forecasting, as shown in Table~\ref{P-values and treatment affects of causality for short-term forecasting}, and the causality effect $\tau$ is around 10\%.
The pandemic information positively impacts the electricity load in the second hour of the next day.
News concerning the politics of Northern Ireland can also positively impact the load demand.
Grid operators would benefit from paying attention to this information when forecasting short-term national load demand.

\subsection{Considerations}
The experiments in this study show that text information significantly improves load demand forecasting.
However, there is still room for discussion about the textual features in the Granger test, Pearson correlations, and the causality effects.
For instance, \emph{Dim-9} is summarized as weapons related to tensions in the Middle East and the assumed consequences on the oil market.
It passed the Granger test, but with a low negative correlation with demand, shown in Figure~\ref{Pearson_analysis_NLP}.
However, the results shown in Figure~\ref{Scatter plot for causality} identify a relatively strong causality effect.
Another example is \emph{Topic-18}, the COVID-19-related topic, which has a higher correlation but cannot cause a change in demand.
The third example is \emph{Dim-69}, the military-related dimension.
Figure~\ref{Pearson_analysis_NLP} does not show that this correlates with demand, and there is no evidence that it can cause demand.
The three examples demonstrate that i) the Granger test can be used as a feature selection technique with low explainability of causality; ii) there is no clear link between Pearson correlation and causality effect, and higher correlation does not mean a higher causality effect.

The work also has some limitations.
Firstly, the BBC website was the only text source, and although it features around one million news items, it would still be worth adding news from other sources.
Secondly, the work was carried out on national aggregated electric demand with a relatively stable pattern. 
Therefore, day-ahead forecast errors are low, and it is difficult to identify the effect of the additional textual features.
Thirdly, the paper contains the basic NLP methods for extracting textual features. More advanced algorithms should be considered.

\section{Conclusions}\label{Conclusions}
This paper studied the link between unstructured textual information in news and electricity demand in the UK.
In general, improvements of 4\%, 11\%, and 10\% in RMSE, MAE, and SMAPE are observed. 
The values should be considered in light of the relatively smooth behavior of the national aggregated load and its predictability.
The best-performing method is a feature combination model with word frequency from news titles, sentiment scores, and GloVe word embeddings from news text bodies.
These features identified keywords relative to sports stars, the COVID-19 pandemic, the minimum subjectivity of public sentiments, transportation, and international conflicts.
The experiments on short-term and regional forecasting demonstrate the potential of textual features.

The contributions listed in Section~\ref{Introduction} are now specified.
First, we verified the possibility of extracting valuable information from the news to improve demand forecasting. 
Table~\ref{Forecasting errors when adding Ft into model D+C+T} shows the improvements in the method when adding textual features.
In Figure~\ref{Pearson_analysis_NLP}, the correlation of textual features and load at different hours, seasons, and day types can be observed.

The second contribution is the prediction chain integrating textual information. 
i) Textual information in the news is converted into numerical time series, including count features, word frequencies, sentiment scores, topic distributions, and word embeddings according to different NLP techniques. 
ii) After the Granger test to remove spurious correlations, the rest of the features are fed to an existing forecasting model with known predictors such as demand lags, calendar information, and temperature values. 
iii) The performance is compared, and the inputs are analyzed to understand the mechanism of the news impacting electricity load from global and local relationships, and causality effects.

The third contribution is that we propose explanations for the mechanism behind forecasting improvement. 
Our key findings are:
i) The effect of the COVID-19 pandemic appears as keywords. 
This mechanism may be related to the reduced demand due to lockdowns enforced in the years 2020-21, causing obvious negative correlations.
ii) News related to Northern Irish politics impacts the electricity demand with a symptom of more generic political instability that causes a demand reduction. 
iii) It is assumed that international tension in the oil-rich region may impact the economy and hence the electric load.

This study, far from closing the subject, opens a new series of questions for further research.
Firstly, the consequences of social sciences, economics, or energy policy findings are potential elements that impact demand forecasting.
Since some social events have a more prolonged impact, it would be preferable to replicate the study on longer horizons.
Secondly, the analysis should be conducted at a higher granularity spatial node level, including the error improvement distribution brought by additional features among all the nodes in the system.
Other methods for text analysis are worth trying, for example, to test n-grams instead of single keywords, and to use more complex deep networks.
Moreover, social media should be explored as additional data sources, including analysis of the multilingual contents or tweets with geographical information.
Finally, probabilistic forecasting is another research scenario to compare our proposed approach against other metrics and challenges, such as extreme loads forecasting.

\bibliographystyle{IEEEtran}
\bibliography{References}

\begin{IEEEbiography}[{\includegraphics[width=1in,height=1.25in,clip,keepaspectratio]{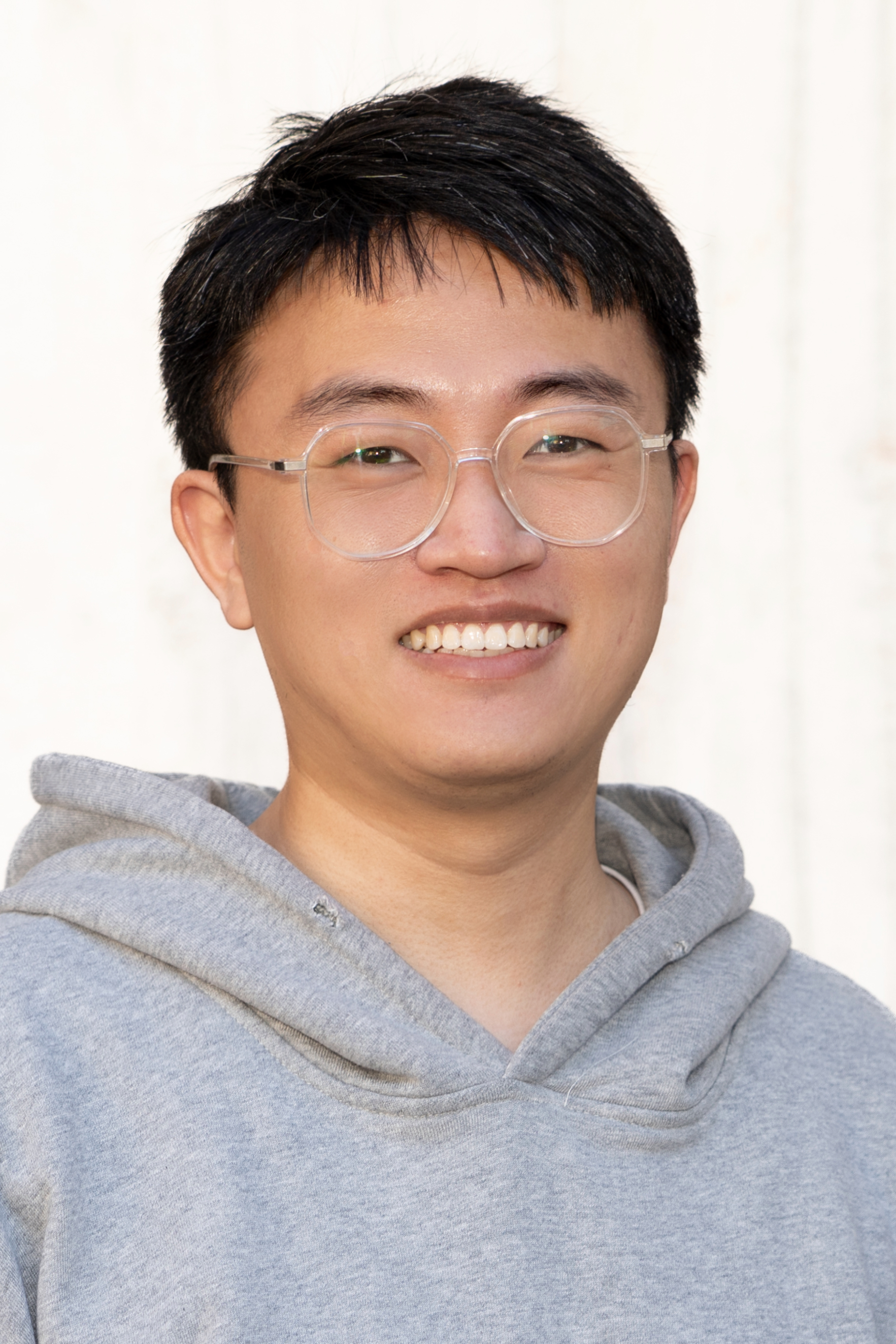}}]{Yun Bai} (Member, IEEE) received a BSc degree in Statistics and an MSc degree in Management Science and Engineering from Beihang University, Beijing, China.
Since 2022, he has been a doctoral student at MINES Paris - PSL University with a scholarship from the China Scholarship Council (CSC). His research interests include energy forecasting, natural language processing applications, and machine learning techniques.
\end{IEEEbiography}
\vspace{-1mm}

\begin{IEEEbiography}[{\includegraphics[width=1in,height=1.25in,clip,keepaspectratio]{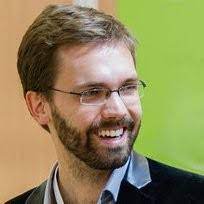}}]{Simon Camal} (Member, IEEE) received a BEng. degree in Energy and Environmental Engineering from Mines Nancy, Nancy, France, in 2010, a European MSc in Renewable Energy from Loughborough University, Loughborough, UK, in 2011, and a PhD from MINES Paris - PSL University, Paris, France, in 2020, on forecasting and optimization of ancillary service provision by renewable energy power plants. He currently works at the MINES ParisTech Center for Processes, Renewable Energies and Energy Systems (PERSEE), Sophia Antipolis, France, as the Project Manager of the Horizon2020 Smart4RES Project.
\end{IEEEbiography}
\vspace{-1mm}

\begin{IEEEbiography}[{\includegraphics[width=1in,height=1.25in,clip,keepaspectratio]{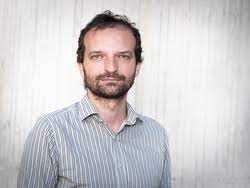}}]{Andrea Michiorri} MSc, PhD, HDR, is a Researcher at the PERSEE Centre of Mines Paris – PSL where he leads research on the integration of renewables into the power systems. 
\end{IEEEbiography}
\end{document}